\documentclass{ws-procs11x85}
\usepackage{ws-procs-thm}
\usepackage[colorlinks=true,linkcolor=blue,urlcolor=blue]{hyperref}
\usepackage[caption=false]{subfig}
\begin{document}





\title{From Detection to Mitigation: Addressing Bias in Deep Learning Models for Chest X-Ray Diagnosis \footnote{Preprint of an article published in Pacific Symposium on Biocomputing © 2026 World Scientific Publishing Co., Singapore, http://psb.stanford.edu/}}

\author{Clemence Mottez$^1$, Louisa Fay$^2$, Maya Varma$^1$, Sophie Ostmeier$^1$, Curtis Langlotz$^1$}

\address{$^1$Center for Artificial Intelligence in Medicine and Imaging, Stanford University, CA\\ $^2$Medical Image and Data Analysis (midas.lab), University Hospital of Tübingen, Germany \\
Emails: cmottez@stanford.edu, louisa.fay@med.uni-tuebingen.de, mvarma2@stanford.edu, sostm@stanford.edu, langlotz@stanford.edu}

\begin{abstract}
Deep learning models have shown promise in improving diagnostic accuracy from chest X-rays, but they also risk perpetuating healthcare disparities when performance varies across demographic groups. In this work, we present a comprehensive bias detection and mitigation framework targeting sex, age, and race-based disparities when performing diagnostic tasks with chest X-rays. We extend a recent CNN–XGBoost pipeline to support multi-label classification and evaluate its performance across four medical conditions. We show that replacing the final layer of CNN with an eXtreme Gradient Boosting classifier improves the fairness of the subgroup while maintaining or improving the overall predictive performance. To validate its generalizability, we apply the method to different backbones, namely DenseNet-121 and ResNet-50, and achieve similarly strong performance and fairness outcomes, confirming its model-agnostic design. We further compare this lightweight adapter training method with traditional full-model training bias mitigation techniques, including adversarial training, reweighting, data augmentation, and active learning, and find that our approach offers competitive or superior bias reduction at a fraction of the computational cost. Finally, we show that combining eXtreme Gradient Boosting retraining with active learning yields the largest reduction in bias across all demographic subgroups, both in and out of distribution on the CheXpert and MIMIC datasets, establishing a practical and effective path toward equitable deep learning deployment in clinical radiology.
\end{abstract}


\copyrightinfo{\copyright\ 2025 The Authors. Open Access chapter published by World Scientific Publishing Company and distributed under the terms of the Creative Commons Attribution Non-Commercial (CC BY-NC) 4.0 License.}

\section{Introduction}

Deep learning (DL) models have demonstrated remarkable success in medical imaging tasks, including disease detection from chest X-rays (CXRs)\cite{chexnet}. These models promise to improve clinical workflows by improving diagnostic accuracy, enabling faster decision-making, and expanding access to care. However, as DL systems become increasingly integrated into healthcare, concerns have emerged about their potential to exacerbate health disparities. In particular, models trained on unbalanced datasets can exhibit biased performance across subgroups defined by sex, age, or race, raising critical issues of fairness, trust, and safety in clinical deployment\cite{yang, Wiens}.

Bias in DL models can come from multiple sources, including underrepresentation in training data, spurious correlations, and learned shortcut features\cite{yang}. These biases may result in systematically worse performance for specific demographic groups, which undermines the equity and reliability of medical AI systems. Existing bias mitigation techniques, such as reweighting samples, adversarial training, and data augmentation, can be effective but often require full model retraining\cite{existing_bias_mit}. This makes them computationally expensive and difficult to implement in real-world healthcare settings, where data access and training resources are limited.

To address these limitations, we propose a lightweight and effective bias mitigation strategy building upon prior work\cite{clem}. The idea is to extract the last layer of a convolutional neural network (CNN), freeze the embeddings, and retrain the head using an eXtreme Gradient Boosting (XGBoost)\cite{XGBoost} classifier. 
Our contributions are as follows:

\begin{itemize}
    \item We perform a detailed bias detection analysis to quantify disparities across sex, age, and race subgroups using large-scale public datasets (CheXpert and MIMIC).
    \item We introduce a CNN-XGBoost pipeline that supports multi-label disease prediction. 
    \item We demonstrate that our XGBoost adapter head can be effectively integrated with different CNN-based architectures, showing similar improvements in performance and reductions in bias. 
    \item We benchmark our method to multiple adaptation heads and to full model fine-tuning and demonstrate that XGBoost offers the best trade-off between performance, fairness, and computational cost.
    \item Finally, we demonstrate that combining XGBoost retraining with active learning is a successful bias mitigation that generalizes to both In-Distribution (ID) and Out-Of-Distribution (OOD) settings.

\end{itemize}
This work presents a practical and scalable path for deploying fair and effective DL models in clinical radiology, enabling safer and more equitable care for diverse patient populations.

\section{Related Work}
Bias in medical Artificial Intelligence (AI) has become a major concern, as DL models may perform unevenly across patient subgroups. Previous research has shown that CXR diagnostic models often encode demographic information, such as sex, age, or race, even when not explicitly trained to do so\cite{existing_bias_mit, yang}. This can lead to biased predictions, especially in the presence of demographic imbalances in the training data.

To address these issues, various fairness-focused methods have been proposed, including sample reweighting, adversarial training, and fairness-aware objectives\cite{existing_bias_mit}. Other studies have explored last-layer retraining to mitigate spurious correlations, showing that simple linear-head replacements can achieve fairness with low computational cost\cite{lastlayer}. 

Some research has investigated CNN–XGBoost hybrid models for performance enhancement, particularly in tasks such as pneumonia or breast cancer detection\cite{Hedhoud2023, article}, but their role in bias mitigation remains underexplored.

A recent study proposed a lightweight bias mitigation strategy that replaces the final layer of a CNN with an XGBoost classifier trained on a curated subset of embeddings\cite{clem}. While this method showed promise in reducing bias related to a single disease, it did not evaluate its applicability to multiple conditions, nor did it compare XGBoost to other classifier types or integrate existing bias mitigation techniques.

In this work, we extend the CNN–XGBoost approach to address its current limitations.

\section{Method and Materials}

\subsection{Data}
We consider two large publicly available CXR datasets in this work:
\vspace{-5pt}
\begin{itemize}
\item \textbf{CheXpert Plus} \cite{CheXpertPlus}: This dataset includes 224,316 CXR images collected at Stanford Health Care and a test set of 500 exams with annotations from eight radiologists.
\item \textbf{Medical Information Mart for Intensive Care (MIMIC)} \cite{MIMIC}: This dataset contains 377,110 CXR images acquired at the Beth Israel Deaconess Medical Center.
\end{itemize}
\vspace{-5pt}
Both datasets include demographic information of sex, age, and race. For sex-based analysis, we compare model performance between male and female patients. For age, we apply a threshold of 70 years to separate the data into two categories: young and old. For race, we focus on three groups: White, Black, and Asian. Patients from other racial backgrounds were grouped under an "Other" category, which was too small to support reliable subgroup analysis.

We follow the training and test splits provided by CheXpert and MIMIC, using only posterior-anterior (PA) and anterior-posterior (AP) view images. This filtering results in 112,105 CXRs for CheXpert and 139,508 for MIMIC. All images are resized to 224×224 or 512x512 pixels to match the input requirements of the respective models.

\subsection{Metrics}
Defining clinical bias is inherently complex, as it involves trade-offs between fairness and other key performance metrics. There is no evidence suggesting that CXRs from certain demographic subgroups are more difficult to classify, and we argue that this parity should be reflected in model behavior as well. In this study, we define bias as disparities in model performance across subgroups. Reducing bias should not come at the cost of overall model performance. However, in some cases, mitigating bias may unintentionally disadvantage specific subgroups. 

To measure overall performance, we use the Area Under the Precision-Recall Curve (AUPRC), which provides a balanced assessment of precision and recall and is particularly suited for imbalanced datasets. To evaluate fairness, we compute the performance disparity across subgroups using $\Delta$AUPRC, the absolute difference in AUPRC between subgroups. In cases with more than two subgroups, such as race, we report the maximum observed $\Delta$AUPRC as the fairness metric. In our framework, the goal is to achieve high AUPRC (strong overall performance) and low $\Delta$AUPRC (minimal disparity across subgroups).

\subsection{Bias Detection Framework}
Before mitigating the bias, it is essential to detect and understand its sources. Model bias can come from various factors, including data composition, clinical context, and the algorithm itself. In this analysis, we focus on identifying potential sources of bias on CheXpert.

\begin{enumerate}
\item \textbf{Data and Clinical Context:}
We study disparities introduced during data collection. For each disease and demographic subgroup, we analyze the distribution of positive and negative labels to assess imbalances. Additionally, we study differences in disease prevalence across demographic groups to understand potential confounding clinical patterns.
\item \textbf{Model Behavior:}  
To investigate the model’s internal representations, we visualize learned embeddings using Principal Component Analysis (PCA) and t-distributed Stochastic Neighbor Embedding (t-SNE)\cite{tSNE} plots, stratified by demographic group. We also evaluate whether demographic attributes (sex, age, race) can be predicted from these embeddings (originally trained to classify medical conditions) to assess whether the model encodes sensitive information. Furthermore, we employ SHapley Additive exPlanations (SHAP) values to analyze feature importance when predicting specific diseases across different subgroups, helping us determine whether the model attends to different features depending on the subgroup. Finally, we assess performance disparities by subgroup. 

\item \textbf{Radiologist Agreement:}  
Since our models are trained on radiologist-generated labels, we assess potential inter-radiologist variability across demographic subgroups. To perform this analysis, we analyze agreement among the 500 predictions of eight radiologists. 
\end{enumerate}


\subsection{Bias Mitigation Methods}

\begin{figure*}[h]
\begin{center}
    \includegraphics[width=1\linewidth]{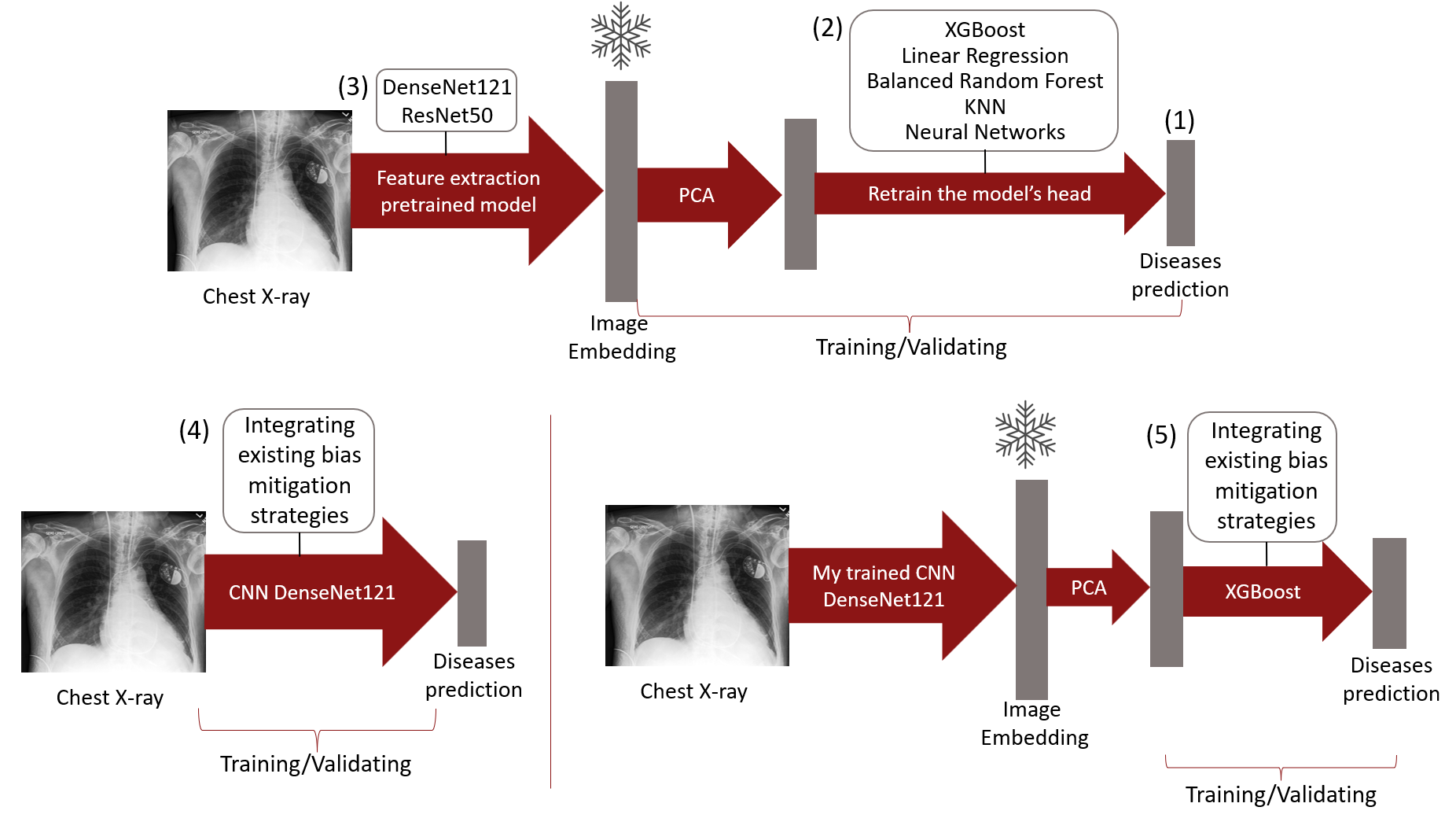}
\end{center}
\vspace{-15pt}
\caption{Bias mitigation method pipeline. The numbers correspond to the respective steps.}
\label{fig:pipeline}
\end{figure*}

To address limitations in the CNN–XGBoost pipeline, we propose several enhancements. The overall pipeline is illustrated in Figure~\ref{fig:pipeline}. The baseline method operates as follows: for each image, a 1024-dimensional feature vector is extracted from the last hidden layer of a pretrained DenseNet-121. This embedding is then reduced using PCA, retaining 95\% of the variance. The resulting lower-dimensional representation is used as input to an XGBoost classifier trained to predict a medical condition. This model is chosen for its ensemble learning capabilities, which enable it to focus on difficult-to-classify instances by iteratively correcting errors made by previous trees. It also handles imbalanced data effectively, an important property given the imbalance both in disease prevalence and across demographic subgroups. The DenseNet-121 model used is the pretrained version from the TorchXRayVision library \cite{torchxrayvision}. It is pretrained on the CheXpert dataset and is evaluated on both CheXpert and MIMIC to ensure generalization.

\begin{enumerate}
\item \textbf{Extension to Multiple Medical Conditions:}
To generalize the method to multiple pathologies, we replace the single-output XGBoost classifier with a multi-head classifier, where each head corresponds to one medical condition. The conditions are selected based on two criteria: (i) the model’s AUC for each condition exceeds 70\%, and (ii) the condition has a at least 10\% of positive sample, ensuring sufficient representation across subgroups.

\item \textbf{Alternative Classifier Heads:}  
To explore the impact of different classifiers on fairness, we replace the XGBoost head with various models using a Multi-OutputClassifier framework, including Logistic Regression (LR), Decision Tree (DT), Random Forest (RF), Neural Network (NN), K-Nearest Neighbors (KNN), and Balanced Random Forest (BRF).

\item \textbf{Model-Agnostic Generalization:}
Because our approach relies on embeddings extracted from image encoders, it is inherently model-agnostic. To test this property, we applied the same pipeline using a ResNet-50 model from the TorchXRayVision library \cite{torchxrayvision}.

\item \textbf{Comparison with Standard Bias Mitigation Techniques:}  For a more controlled setting, we retrain a DenseNet-121 from scratch. This allows us to control the dataset splits and tailor the output layer to predict only the medical conditions relevant to our analysis. For initialization, we use pretrained weights from CheXNet \cite{chexnet}, a DenseNet-121 model trained specifically for pneumonia detection (note that pneumonia is not among the target conditions in our study). The training configuration follows the defaults used in TorchXRayVision \cite{torchxrayvision}. Then we apply several existing bias mitigation strategies and compare their effectiveness to that of XGBoost-based head retraining:
\begin{itemize}
    \item \textbf{Weighted Sampling:} Reweighting the training data to balance subgroups.  
    \item \textbf{Adversarial Training:} Introducing a secondary adversarial branch to predict sensitive attributes (sex, race, age), while training the primary network to be demographically agnostic by minimizing this branch’s accuracy.  
    \item \textbf{Targeted Data Augmentation:} Augmenting under-performing subgroups. 
    \item \textbf{Active Learning:} Prioritizing the inclusion of underrepresented or uncertain samples via uncertainty or diversity-based sampling.
\end{itemize}

\item \textbf{Combining Bias Mitigation Strategies:}  
Finally, we combine the above mitigation techniques with XGBoost head retraining. This hybrid approach allows us to benefit from XGBoost's ability to handle imbalanced feature distributions, while simultaneously correcting for bias embedded in the data. This strategy is computationally more efficient than full model retraining.
\end{enumerate}

For hyperparameter tuning, we use a custom score function designed to balance performance and fairness:
$Score = AUPRC - (\Delta AUPRC_{sex} + \Delta AUPRC_{age} + \Delta AUPRC_{race})$. This formulation encourages the model to achieve high overall predictive performance while minimizing disparities across age, race, and sex.
Hyperparameters were selected based on the model performance evaluated on the validation dataset. Each experiment is run five times, the results are averaged and the standard deviation and CI are computed.

\section{Experiments, Results, and Discussion}
\subsection{Bias Detection Analysis}
To better understand the origins of bias in our model, we analyzed three potential sources: the data distribution, the model’s internal representations, and human (radiologist) variability.

\begin{enumerate}
\item \textbf{Data and Label Distribution:}
    We first studied the class distribution across demographic subgroups in the CheXpert dataset (Table~\ref{tab:demo_subgroup}). Several imbalances were evident:
    \begin{itemize}
        \item Sex: Both males and females exhibited similar prevalence rates across conditions. For example, Lung Opacity appeared in 49.5\% of both groups. 
        \item Age: Older patients showed significantly higher disease prevalence. For instance, Cardiomegaly was present in 15.4\% of older patients versus 10.3\% in younger ones. This trend was consistent across all four diseases studied. 
        \item Race: Substantial variation was observed in disease prevalence. For example, Cardiomegaly prevalence in Black patients was 18.3\%, higher than in White (11.5\%) or Asian (12.9\%) patients. Moreover, there is a high data imbalance according to race, where White people represent 78.2\% of the data, Asian 14.7\% and Black 7.1\%.
        \end{itemize}

\begin{table}[h]
\centering
\begin{tabular}{ll|rrr|rrr|rrr|rrr}
\multicolumn{2}{c|}{} 
  & \multicolumn{3}{c|}{Cardiomegaly} 
  & \multicolumn{3}{c|}{Lung Opacity} 
  & \multicolumn{3}{c|}{Edema} 
  & \multicolumn{3}{c}{Pleural Effusion} \\
 &  
  & \(-\) & \(+\) & \(\%\!+\) 
  & \(-\) & \(+\) & \(\%\!+\) 
  & \(-\) & \(+\) & \(\%\!+\) 
  & \(-\) & \(+\) & \(\%\!+\) \\

Sex   & Female & 37.3 &  4.7 & 11.2 & 21.2 & 20.8 & 49.5 & 31.9 & 10.1 & 24.0 & 25.3 & 16.7 & 39.8 \\   & Male   & 50.5 &  7.5 & 12.9 & 29.3 & 28.7 & 49.5 & 44.3 & 13.7 & 23.6 & 34.9 & 23.1 & 39.8 \\
\hline
Age   & Young  & 56.4 &  6.5 & 10.3 & 33.2 & 29.7 & 47.2 & 49.5 & 13.4 & 21.3 & 39.4 & 23.5 & 37.4 \\
& Old    & 31.4 &  5.7 & 15.4 & 17.2 & 19.9 & 53.6 & 26.7 & 10.4 & 28.0 & 20.9 & 16.2 & 43.7 \\
\hline
Race  & White  & 69.2 &  9.0 & 11.5 & 39.3 & 38.9 & 49.7 & 59.4 & 18.8 & 24.0 & 46.9 & 31.3 & 40.0 \\
  & Asian  & 12.8 &  1.9 & 12.9 &  7.5 &  7.2 & 49.0 & 11.5 &  3.1 & 21.2 &  8.6 &  6.1 & 41.5 \\
  & Black  &  5.8 &  1.3 & 18.3 &  3.7 &  3.4 & 47.9 &  5.3 &  1.9 & 26.4 &  4.8 &  2.4 & 33.3 \\
\end{tabular}
\caption{CheXpert data class distribution across demographic subgroups and diseases. For each disease studied, and for each subgroup, we analyze the percentage of negative samples (–), positive samples (+) and among the specific subgroup the percentage of positive labels in comparison to negative ones (\(\%\!+\)).}
\label{tab:demo_subgroup}

\end{table}
These disparities could introduce confounding if not accounted for. 

\item \textbf{Model Representations and Learned Biases:} 
To investigate whether the model encodes demographic information implicitly, we first visualized the extracted embeddings using PCA and t-SNE. As shown in Figure~\ref{fig:emb}, we observe visual patterns that suggest a degree of separation across sex, age, and race subgroups, indicating the presence of demographic signals in the learned representations.

\begin{figure*}[h]
\begin{center}
    \includegraphics[width=0.8\linewidth]{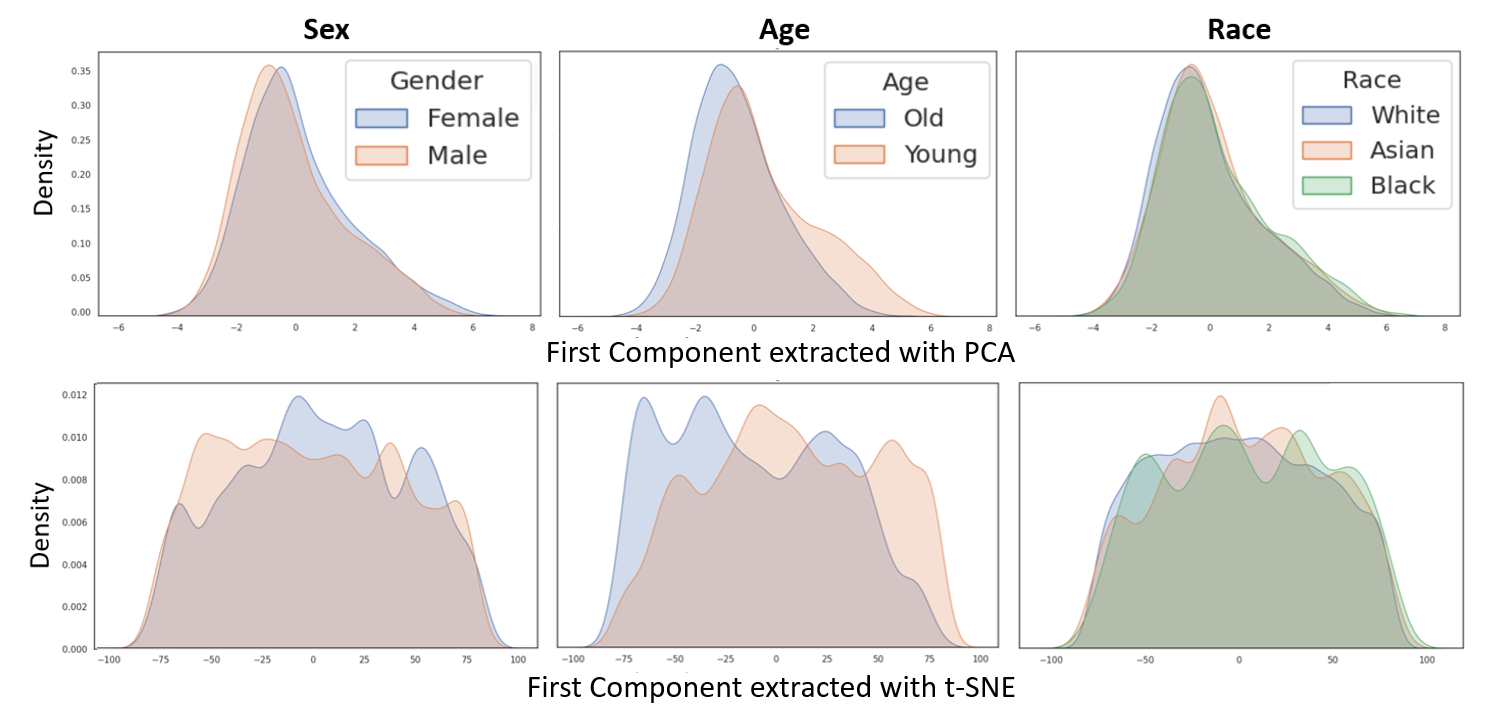}
\end{center}
\vspace{-10pt}
\caption{PCA and t-SNE first components of the extracted embeddings according to the different demographic subgroups.}
\label{fig:emb}
\end{figure*}

To quantify this observation, we trained a simple LR classifier to predict demographics from CNN embeddings. It achieved high AUCs (sex: 0.93, age: 0.82, race: 0.77), confirming that the model encodes demographic information despite not being trained to do so.

Further, we used SHAP to examine whether the model's feature attributions differ across demographic subgroups when predicting specific medical conditions. Specifically, we analyzed whether the most influential embedding dimensions contributed in a consistent direction across subgroups when predicting Cardiomegaly. Table~\ref{tab:direction-consistency} presents the direction analysis for the five most important embedding dimensions. For each embedding and subgroup, we indicate whether the SHAP value direction was consistent (“same”) or reversed (“opposite”) between subgroups. For example, Embedding 950 showed opposite attributions across all three demographics, suggesting the model interprets this feature differently depending on the patient's demographics. Such representational differences point to learned bias, which becomes problematic when linked to performance disparities.

\begin{table}[ht]
\centering
\begin{tabular}{c|ccc}
Embeddings & Sex & Age & Race \\
\hline
773 & same      & same      & same      \\
781 & opposite  & same      & same      \\
950 & opposite  & opposite  & opposite  \\
696 & same      & opposite  & opposite  \\
603 & same      & opposite  & same      \\
\end{tabular}
\caption{Direction consistency of SHAP values across subgroups (sex, age, race) for Cardiomegaly prediction. Only the five most influential embeddings are shown.}
\label{tab:direction-consistency}
\end{table}

To assess this, we computed the mean $\Delta$AUPRC across the four medical conditions. We found persistent bias: $\Delta$AUPRC of 1.6 for sex, 4.1 for age, and 8.7 for race. These values indicate that while the model may achieve high overall accuracy, its performance is not distributed equally across demographic groups, especially with respect to race. This underscores the importance of addressing representational and predictive disparities through targeted bias mitigation strategies.

\item \textbf{Radiologist Variability:}
We analyzed inter-rater disagreement among eight radiologists on 500 patients (Table~\ref{tab:counts}). While radiologists can sometimes visually infer sex and roughly estimate age, they cannot identify a patient’s race from a CXR\cite{race}. Therefore, any racial bias observed in the model is more likely to come from the data or algorithm. 
We computed disagreement rates as the average proportion of radiologists who did not agree with the majority vote for each case, stratified by subgroup. Disagreements, as shown in Table~\ref{tab:disagreement}, were generally low and consistent, with a bigger difference related to the age subgroup, most likely due to disease prevalence in older patient. 
This suggests that radiologist uncertainty does not disproportionately affect any subgroup, meaning label noise is likely not a major contributor to subgroup bias.

\begin{table}[ht]
\centering
\begin{minipage}[t]{0.45\textwidth}
  \centering
  \begin{tabular}{ll|c}
    & Subgroup     & Count \\
    \hline
    Sex & Male     & 202 \\
    & Female   & 125 \\
    \hline
    Age& Young    & 196 \\
    & Old      & 131 \\
    \hline
    Race &White   & 276 \\
    & Asian   &  51 \\
  \end{tabular}
  \caption{Counts by subgroup}
  \label{tab:counts}
\end{minipage}\hfill
\begin{minipage}[t]{0.45\textwidth}
  \centering
  \begin{tabular}{ll|c}
    & Subgroup     & \% Disagreement \\
    \hline
    Sex & Male     & 9.3 \\
    & Female   & 9.6 \\
    \hline
    Age &Young    & 8.8 \\
    & Old      & 10.0  \\
    \hline
    Race &White   & 9.6 \\
    & Asian   & 8.7 \\
  \end{tabular}
   \caption{Disagreement by subgroup}
  \label{tab:disagreement}
\end{minipage}
\end{table}

\end{enumerate}

\subsection{Bias Mitigation Analysis}
We evaluate several strategies to mitigate bias while maintaining strong performance. Our approach builds upon the CNN–XGBoost pipeline by progressively enhancing its flexibility, evaluating its robustness, and benchmarking it against standard bias mitigation techniques.

\begin{figure*}[h]
\begin{center}
    \includegraphics[width=0.98\linewidth]{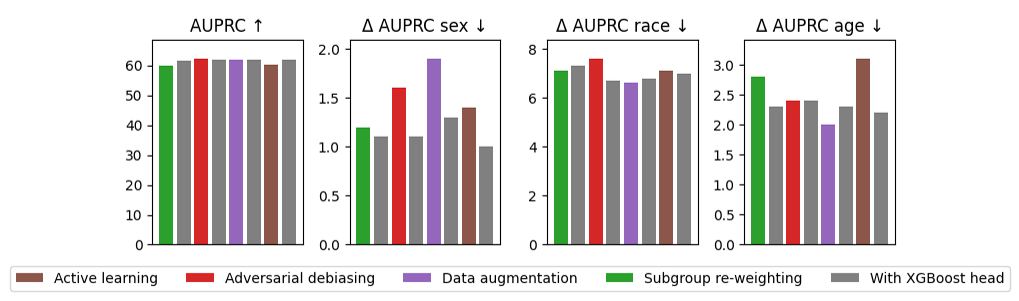}
\end{center}
\vspace{-10pt}
\caption{Comparison of existing bias mitigation methods that require full model retraining with our method combined with our XGBoost head retraining (in grey).}
\label{fig:compare}
\end{figure*}

\begin{figure*}[h]
\begin{center}
    \includegraphics[width=0.68\linewidth]{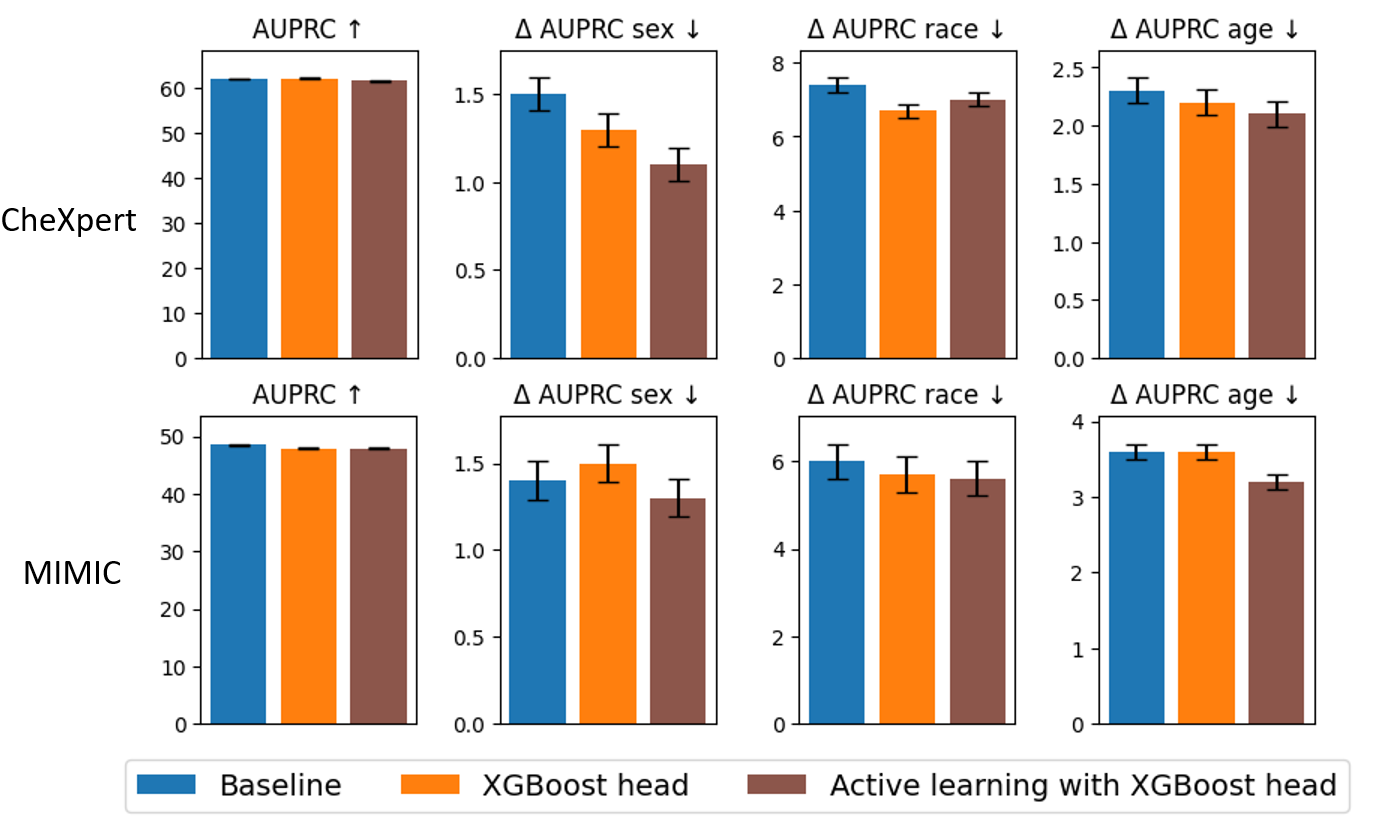}
\end{center}
\vspace{-10pt}
\caption{Comparison of initial performance and bias (baseline in blue) with XGBoost head retraining (orange), and XGBoost head retraining combined with active learning (brown), ID on CheXpert and OOD on MIMIC. }
\label{fig:final}
\end{figure*}

\begin{enumerate}
    \item \textbf{Extension to Multi-Label Classification:}
We first extended the baseline method, which focused only on Pleural Effusion \cite{clem}, to handle multiple medical conditions. Specifically, we added Cardiomegaly, Lung Opacity, and Edema in the analysis. As shown in Figure~\ref{fig:baseline}, the multi-label extension improves overall performance while reducing bias.

\begin{figure*}[h]
\begin{center}
    \includegraphics[width=0.5\linewidth]{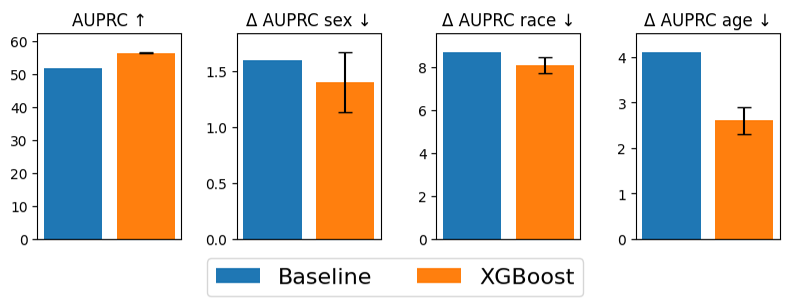}
    \label{fig:img1}
\end{center}
\vspace{-10pt}
\caption{Performance and bias between the baseline DenseNet-121 model and the model with the head retrained with an XGBoost classifier on CheXpert.}
\label{fig:baseline}
\end{figure*}

\item \textbf{Evaluation of Alternative Classifier Heads:}
We next evaluated alternative models for the classifier head, replacing XGBoost with LR, DT, RF, NN, BRF, and KNN. As shown in Figure~\ref{fig:models}, XGBoost offered the best trade-off between performance and fairness. LR performed well, which is consistent with the linear structure of the original CNN classifier layer. Notably, LR reduced bias across sex and age but was less effective for race, likely due to higher data imbalance. In contrast, BRF and XGBoost were most robust across all subgroups due to their ensemble design and handling of imbalance. DT and RF performed near random and were excluded. These findings highlight that classifier choice impacts the pattern of bias reduction, with some models more sensitive to subgroup imbalance.

\begin{figure*}[h]
\begin{center}
    \includegraphics[width=0.8\linewidth]{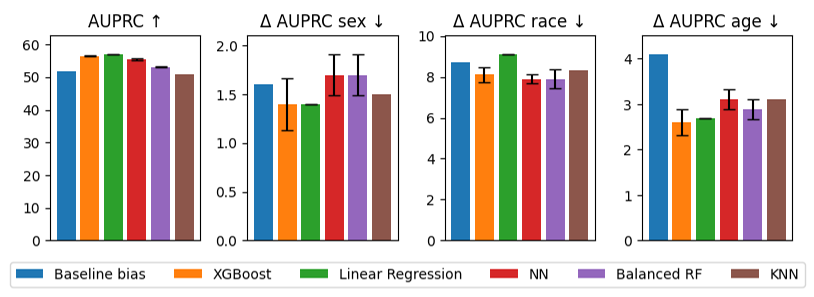}
\end{center}
\vspace{-10pt}
\caption{Differences in performance and bias when retraining the head of the DenseNet-121 with different models on CheXpert. Confidence Intervals (CI) are not shown for KNN since it doesn't involve any internal randomness or stochastic training process.}
\label{fig:models}
\end{figure*}

\item \textbf{Generalization to Other Backbone Architectures:}
To evaluate the model-agnostic nature of our framework, we repeated our experiments using a ResNet-50 architecture with a 512-dimensional embedding output. As with DenseNet-121, we first applied PCA and then retrained the final head using XGBoost. The results presented in Figure \ref{fig:resnet} mirrored those observed with DenseNet: we achieved a consistent increase in overall performance and a noticeable reduction in bias across sex, age, and race subgroups. This confirms that the bias mitigation approach can be flexibly applied across different CNN backbones. 


\begin{figure*}[h]
\begin{center}
    \includegraphics[width=0.5\linewidth]{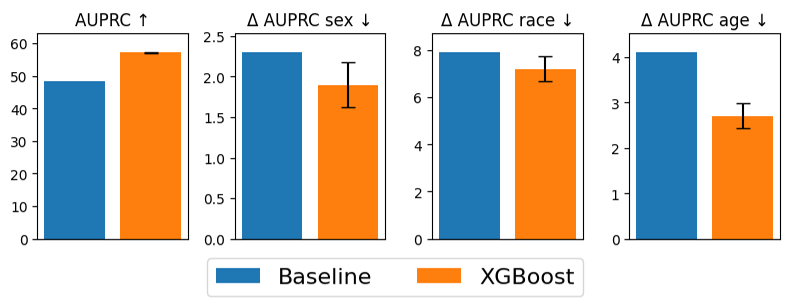}
\vspace{-5pt}
\caption{Performance and bias between the ResNet-50 model and the model with the head retrained with an XGBoost on CheXpert.}
\label{fig:resnet}
\end{center}
\end{figure*}

\item \textbf{Full Model Retraining versus Lightweight Head Retraining:}
We retrained a DenseNet-121 CNN from scratch for fairer comparisons. We then compared our lightweight XGBoost method with existing bias mitigation approaches that require full model retraining. As shown in Figure~\ref{fig:full}, XGBoost head retraining achieved comparable or even superior performance in reducing bias, at a fraction of the computational cost. Specifically, our method retrains only $\sim$20,000 parameters, versus $\sim$8 million in a full DenseNet-121.

\begin{figure*}[h]
\begin{center}
    \includegraphics[width=0.98\linewidth]{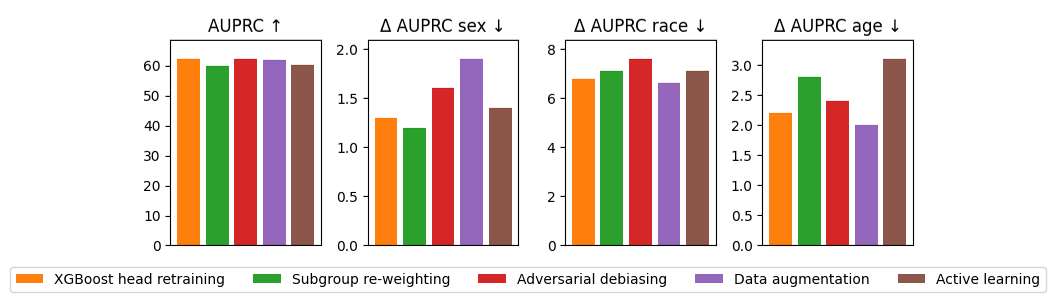}
\vspace{-5pt}
\caption{Comparison of our lightweight bias mitigation method (in orange) with existing methods that require full model retraining.}
\label{fig:full}
\end{center}
\end{figure*}

\item \textbf{Combining Bias Mitigation Techniques:}
Finally, we combined XGBoost head retraining with standard bias mitigation strategies, such as weighted sampling, adversarial training, data augmentation, and active learning, and compared the results with applying these strategies to full model retraining. As illustrated in Figure~\ref{fig:compare}, combining mitigation strategies with XGBoost retraining consistently outperformed full model retraining, both in performance and fairness, and again at much lower computational cost.

The final results, presented in Figure~\ref{fig:final}, show that the combination of active learning with XGBoost head retraining yields the largest reduction in bias across all subgroups sex, age, and race, both ID on CheXpert and OOD on MIMIC. The optimized hyperparameters for XGBoost are as follows: \texttt{eval\_metric = 'logloss'}, \texttt{learning\_rate = 0.05}, \texttt{n\_estimators = 150}, and \texttt{max\_depth = 10}. For active learning, we used a pool-based approach starting with 15,000 labeled images and adding 2,000 uncertain samples per round over 10 rounds, for a final training set of 35,000 images.

\end{enumerate}

\vspace{-10pt}

\subsection{Clinical Significance of Bias Mitigation}
While metrics like AUPRC and $\Delta$AUPRC are essential for evaluating model performance and fairness, it is equally important to interpret these results in the context of clinical impact. Our experiments demonstrate that $\Delta$AUPRC can be reduced without sacrificing overall performance. By minimizing performance gaps between sex, age, and race subgroups, we reduce the risk that some populations receive less accurate diagnoses. This helps prevent misdiagnoses in underrepresented groups, which have historically experienced healthcare disparities.

For illustration, we evaluated Pleural Effusion prediction across women of different races. To minimize False Negative Rates (FNR), thresholds were chosen before and after bias mitigation to ensure recall greater than 0.95.
\begin{table}[h!]
\centering
\begin{tabular}{l|ccccccc}
 & FNR White & FNR Asian & FNR Black & $\Delta$FNR & $\Delta$TPR & $\Delta$FPR & EO max gap \\
\hline
Before & 0.159 & 0.136 & 0.154 & 0.023 & 0.023 & 0.015 & 0.023 \\
After  & 0.149 & 0.139 & 0.143 & 0.010 & 0.010 & 0.007 & 0.010 \\
\end{tabular}
\caption{Performance metrics before and after bias mitigation for Pleural Effusion detection in women across racial subgroups.}
\label{tab:impact}
\end{table}

As shown in Table~\ref{tab:impact}, bias mitigation reduced both subgroup FNRs and disparities in Equalized Odds (EO)\cite{eqodds}, with sensitivity and specificity gaps cut roughly in half. Clinically, this means patients receive better and more consistent diagnostic across racial groups, increasing both reliability and trust in the model. Such improvements enhance the likelihood of clinician adoption of AI tools that demonstrate equitable behavior across diverse populations.

\section{Conclusion and Future Work}
In this work, we introduced a practical and lightweight framework for detecting and mitigating demographic bias in DL models for CXR diagnosis. By replacing the final classification layer of a CNN with an XGBoost model, we demonstrated that it is possible to significantly reduce disparities across sex, age, and race subgroups while preserving, if not improving, overall model performance. Our approach generalizes effectively across multiple medical conditions and remains robust in both ID (CheXpert) and OOD (MIMIC) evaluations.

Through our experiments, we showed that:
\begin{itemize}
    \item XGBoost outperforms alternative classifier heads in balancing fairness and accuracy.
    \item The method is model-agnostic and can be applied to any architecture capable of extracting embeddings from images.
    \item Our method rivals or exceeds traditional full-model bias mitigation techniques, including weighted sampling, adversarial training, data augmentation, and active learning, while requiring far fewer computational resources.
    \item Combining our XGBoost head retraining with active learning yields the most substantial bias reduction across all subgroups while maintaining a competitive performance.
\end{itemize}
These findings offer a compelling pathway for deploying fair and efficient medical AI models in real-world clinical settings where computational constraints are often a major barrier.

Despite promising results, this study has several limitations. The racial subgroup analysis is affected by class imbalance, particularly for Black patients. This underrepresentation limits the statistical robustness of bias evaluations and may obscure subtle disparities. Moreover, our work focuses only on CNN-based models applied to CXRs.  Therefore, the generalizability of our findings to other imaging modalities  (e.g., CT, MRI) and tasks (e.g., segmentation) remains to be established. Finally, our approach relies on last-layer retraining. While efficient, it may be insufficient to fully mitigate spurious correlations compared with approaches leveraging representations from all network layers.

Future work include extending the framework to other architectures such as Vision Transformers (ViTs) and applying and validating this approach on other imaging modalities and in tasks beyond classification. 

\section*{Code Availability} Our code is publicly available on our \href{https://github.com/clemence-mottez/Detecting_And_Mitigating_Bias_in_DL_Models_for_Chest_XRay_Diagnosis}{GitHub}.
The repository includes documentation and scripts to adapt our bias detection and mitigation framework to new datasets and tasks.

\section*{Acknowledgment} This work was supported in part by the Medical Imaging and Data Resource Center, which is funded by the National Institute of Biomedical Imaging and Bioengineering under contract 75N92020C00021 and through the Advanced Research Projects Agency for Health.


\newpage
\bibliographystyle{plain}  

\end{document}